\documentclass{article}

\usepackage{proceed2e}
\usepackage{times}
\usepackage{graphicx} 
\usepackage{subfigure}
\usepackage{natbib}
\usepackage{hyperref}
\usepackage{amsmath,amssymb}
\usepackage{color}
\usepackage{array}
\usepackage{multirow}
\usepackage{adjustbox}

\title{Environment-Independent Task Specifications via GLTL}

\author{Michael L. Littman \And Ufuk Topcu \And Jie Fu \And Charles Isbell \And Min Wen \And James MacGlashan} 

\newcommand{\always}{\mbox{$\Box$}}
\newcommand{\eventually}{\diamondsuit}
\newcommand{\until}{\,\mathcal{U}}
\newcommand{\acc}{\mbox{\emph{acc}}}
\newcommand{\rej}{\mbox{\emph{rej}}}

\begin{document}

\maketitle

\begin{abstract}
We propose a new task-specification language for Markov decision
processes that is designed to be an improvement over reward functions
by being environment independent. The language is a variant of Linear
Temporal Logic (LTL) that is extended to probabilistic specifications
in a way that permits approximations to be learned in finite time. We
provide several small environments that demonstrate the advantages of
our geometric LTL (GLTL) language and illustrate how it can be used to
specify standard reinforcement-learning tasks straightforwardly.
\end{abstract}

\section{Introduction}

The thesis of this work is that (1) rewards are an excellent
way of controlling the behavior of agents, but (2) rewards are
difficult to use for specifying behaviors in an
environment-independent way, therefore (3) we need intermediate
representations between behavior specifications and reward functions.


The intermediate representation we propose is a novel variant of
linear temporal logic
that is modified to be probabilistic
so as to better support reinforcement-learning tasks.
Linear temporal
logic has been used in the past to specify reward functions that
depend on temporal sequences~\citep{bacchus96}; here, we expand the
role to provide a robust and consistent semantics that allows desired behaviors
to be specified in an environment-independent way.
Briefly, our approach involves the specification of tasks via temporal
operators that have a constant probability of expiring on each
step. As such, it bears a close relationship to the notion of
discounting in standard Markov decision process (MDP) reward functions~\citep{Puterman94}.

At a philosophical level, we are viewing behavior specification as a
kind of programming problem. That is, if we think of a Markov decision
process (MDP) as an input, a reward function as a program, and a
policy as an output, then reinforcement learning can be viewed as a
process of program interpretation. We would like the same program to
work across all possible inputs.

\subsection{Specifying behavior via reward functions}
\label{s:rf}

An MDP consists of a finite state space, action space,
transition function, and reward function. Given an environment, an
agent should behave in a way that maximizes cumulative discounted
expected reward. The problems of learning and planning in such
environments have been vigorously studied in the AI community for over
25 years~\citep{watkins89,boutilier99,strehl09}. A reinforcement-learning (RL) agent needs
to learn to maximize cumulative discounted expected reward starting
with an incomplete model of the MDP itself.

For ``programming" reinforcement-learning agents, the state of the art
is to define a reward function and then for the learning agent to
interact with the environment to discover ways to maximize its
reward. Reward-based specifications have proven to be extremely valuable for optimal planning in complex, uncertain environments~\citep{russell94}. However, we can show that reward
functions, as they are currently structured, are very difficult to
work with as a way of reliably specifying tasks. The best use case for
reward functions is when the utilities of all actions and outcomes can
be expressed in a consistent unit, for example, time or money or energy. In reality,
however, putting a meaningful dollar figure on scuffing a wall or
dropping a clean fork is challenging. When informally adding negative
rewards to undesirable outcomes, it is difficult to ensure a
consistent semantics over which planning and reasoning can be carried
out.
Further, reward values often need to be changed if the environment itself
changes---they are not environment independent. Therefore, to get a
system to exhibit a desired behavior, it can be necessary to try
different reward structures and carry out learning multiple times in
the target environment, greatly undermining the purpose of autonomous
learning in the first place.


\begin{figure}
\vskip 0.2in
\begin{center}
\centerline{\includegraphics[width=0.2\textwidth]{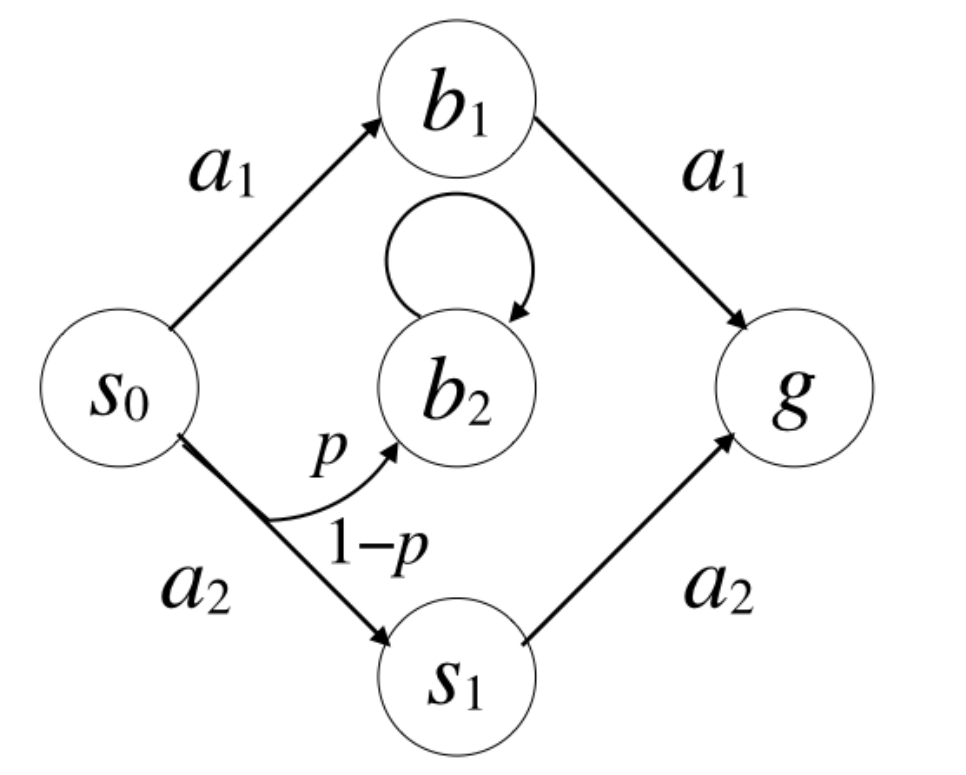}}
\caption{Action $a_2$ has probability $1-p$ of transitioning to a
non-$b$ state and a probability of $p$ of entering a self-loop in a $b$ state. Action $a_1$
passes through a $b$ state and then over to the goal.}
\label{f:nopenalty}
\end{center}
\vskip -0.2in
\end{figure}

Consider the simple example MDP in Figure~\ref{f:nopenalty}.
The agent is choosing between $a_1$ and $a_2$ in the initial state
$s_0$. Choosing $a_1$ causes the agent to pass through \emph{bad}
state $b_1$ for one step, then to continue on to
the goal $g$. Action $a_2$, however, results in a probabilistic transition
to $s_1$ (with probability $1-p$) or bad state $b_2$ (with slip
probability $p$). From $s_1$, the agent can continue on to the
goal. If it reaches $b_2$, it gets stuck there forever.

Let's say our desired behavior is ``maximize the probability of
reaching $g$ without hitting a bad state''. (A bad state could be
something like colliding with a wall or bumping up against a table.) The
probability of success of $a_1$ is zero and $a_2$ is $1-p$. Thus, for
any $0 \le p < 1$, it is better to take action $a_2$.

What reward function encourages this behavior? For concreteness, let's
assume a discount of $\gamma=0.8$ and a reward of $+1$ for reaching the
goal. We can assign bad states a value of $-r$. In the case where
$p=0.1$, setting $r>0.16$ encourages the desired behavior.

Consider, though, what happens if the slip probability is $p=0.3$. Now,
there is no value of $r$ for which $a_2$ is preferred to
$a_1$\footnote{Actually, $r < -24/50$ works for this example, but that
  is tantamount to rewarding the agent for bumping into
  things---something bound to result in other problems.}. That is, it has
become impossible to find a reward function that creates the correct
incentives for the desired behavior to be optimal.



This example is perhaps a bit contrived, but we have observed the same
phenomenon in large and natural state spaces as well. The reason for this result  is that reward functions force us to express utility in terms of
the discounted expected visit frequency of states. In this case, we
are stuck trying to make a tradeoff between the certainty of
encountering a bad state once and the possibility of encountering a
bad state repeatedly. Since we are trying to maximize the
probability of \emph{zero} encounters with a bad state, the expected number
of encounters is only useful for distinguishing zero from more than
zero---the objective cannot be translated into a reward function when
bad states are unavoidable.

\subsection{Specifying behavior via LTL}

An alternative to specifying tasks via reward functions is to use a
formal specification like linear temporal logic or LTL~\citep{pnueli-specification,
modelcheckingbook}.  

Linear temporal logic formulas are built up from a set of atomic propositions; the logic
connectives: negation ($\neg$), disjunction ($\vee$), conjunction
($\wedge$) and material implication ($\rightarrow$); and the temporal
modal operators: next ($\bigcirc$), always ($\always$), eventually
($\eventually$) and until ($\until$). A wide class of properties
including safety ($\always \neg b$), goal guarantee ($\eventually g$),
progress ($\always \eventually g$), response ($\always (b \rightarrow
\eventually g)$), and stability ($\eventually \always g$), where $b$
and $g$ are atomic propositions, can be expressed as LTL
formulas. More complicated specifications can be obtained from the
composition of such simple formulas. For example, the specification of
``repeatedly visit certain locations of interest in a given order
while avoiding certain other unsafe or undesirable locations" can be
obtained through proper composition of simpler safety and progress
formulas~\citep{pnueli-specification, modelcheckingbook}.

Returning to the example in Figure~\ref{f:nopenalty}, the task is to
avoid $b$ states until $g$ is reached: $\neg b \until g$.  Given an
LTL specification and an environment, an agent, for example, should
adopt a behavior that maximizes the probability that the specification is
satisfied. One advantage of this approach is its ability to specify
tasks that cannot be expressed using simple reward functions (like the
example MDP in Section~\ref{s:rf}). Indeed, in the context of
reinforcement-learning problems, we have found it very natural to
express standard MDP task specifications using LTL.

Standard MDP tasks can be expressed well using these temporal
operators. For example:
\begin{itemize}
\item Goal-based tasks like mountain car~\citep{moore91}: If $p$
  represents the attribute of being at the goal (the top of the hill,
  say), $\eventually p$ corresponds to eventually reaching the goal.
\item Avoidance-type tasks like cart pole~\citep{Barto83}: If $q$
represents the attribute of being in the failure state (dropping the
pole, say), $\always \neg q$ corresponds to always avoiding the
failure state.
\item Sequence tasks like taxi~\citep{dietterich00}: If $p$ represents
some task being completed (getting the passenger, say) and $q$
represents another task being completed (delivering the passenger,
say), $\eventually (p \wedge \eventually q)$ corresponds to eventually
completing the first task, then, from there, eventually completing the
second task.
\item Stabilizing tasks like pendulum swing up~\citep{atkeson94}: If
  $p$ represents the property that needs to be stabilized (the
  pendulum being above the vertical, say), $\eventually \always p$
  corresponds to eventually achieving and continually maintaining the
  desired property.
\item Approach-avoid tasks like the 4$\times$3 grid~\citep{russell94}:
If $p$ represents the attribute of being at the goal (the upper right
corner the grid, say), and $q$ represents the attribute of being at a
bad state (the state below it, say), $\neg q \mathcal{U} p$
corresponds to avoiding the bad state en route to the goal.
\end{itemize}

On the other hand, there are barriers to straightforwardly adopting
temporal logic-based languages in a reinforcement-learning setup. The
most significant is that we can show that it is simply impossible to
learn to satisfy classical LTL specifications in some cases. A key
property for being able to learn near-optimal policies efficiently in
the context of reward-based MDPs is what is known as the Simulation
Lemma~\citep{kearns98}. Informally, it says that, for any MDP and any
$\epsilon>0$, there exists an $\epsilon'>0$ such that finding optimal
policies in an $\epsilon'$-close model of the real environment results
in behavior that is $\epsilon$-close to optimal in the real
environment.

Unfortunately, tasks specified via LTL do not have this property. In
particular, there is an MDP and an $\epsilon>0$ such that no
$\epsilon'$-close approximation for $\epsilon'>0$ is sufficient to
produce a policy with $\epsilon$-close satisfaction probability.

\begin{figure}
\vskip 0.2in
\begin{center}
\centerline{\includegraphics[width=0.20\textwidth]{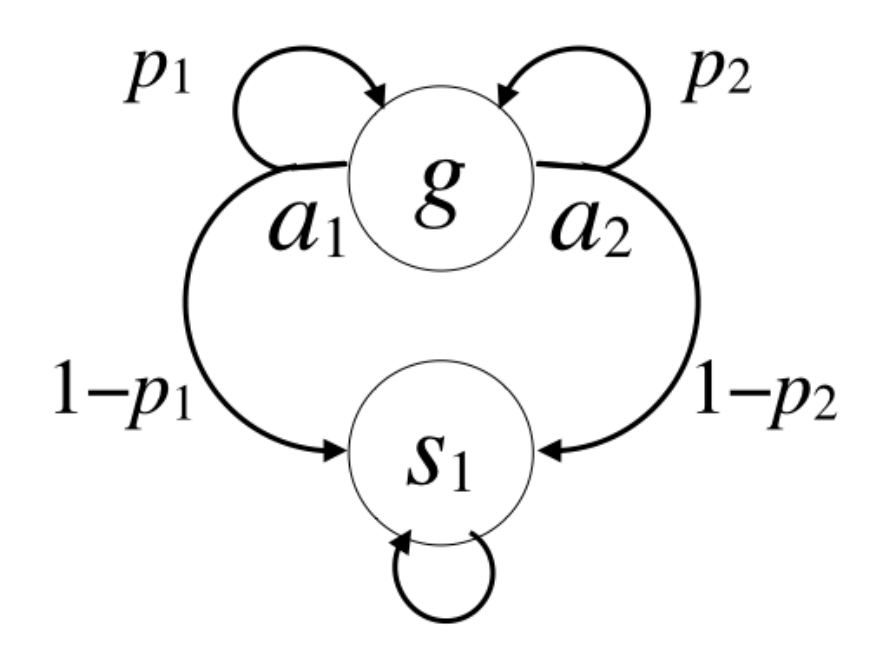}}
\caption{Action $a_1$ has probability $p_1$ of a self-loop and $1-p_1$
  of transitioning to a non-$g$ state. Action $a_2$ has probability
  $p_2$ of a self-loop and $1-p_2$ of transitioning to a non-$g$
  state. The policy that maximizes the probability of satisfaction of
  $\always g$ is highly dependent on $p_1$ and $p_2$ if they are near
  one.} \vspace{-4mm}
\label{f:unlearn}
\end{center}
\vskip -0.2in
\end{figure}
Consider the MDP in Figure~\ref{f:unlearn}. If we want to find a
behavior that nearly maximizes the probability of satisfying the
specification $\always g$ (stay in the good state forever), we need
accurate estimates of $p_1$ and $p_2$. If $p_1 = p_2 = 1$ or $p_1<1$
and $p_2<1$, either policy is equally good. If $p_1=1$ and $p_2<1$,
only action $a_1$ is near optimal. If $p_2=1$ and $p_1<1$, only action
$a_2$ is near optimal. As there is no finite bound on the number of
learning trials needed to distinguish $p_1=1$ from $p_1<1$, a near
optimal behavior cannot be found in worst-case finite time. LTL expressions are
simply too unforgiving to be used with any confidence in a learning
setting.


In this work, we develop a hybrid approach for specifying behavior in
reinforcement learning that combines the strengths of both reward
functions and temporal logic specifications.

%
%


\section{Learning To Satisfy LTL}

While provable guarantees of efficiency and optimality have been at
the core of the literature on
learning~\citep{fiechter94,kearns02,brafman02,li11}, correctness with
respect to complicated, high-level
task specifications---during the learning itself or in the behavior
resulting from the learning phase---has attracted limited
attention~\citep{abbeel05}.

\subsection{Geometric linear temporal logic}

We present a variant of LTL we call geometric linear temporal logic
(GLTL) that builds on the logical and temporal operators in LTL while
ensuring learnability. The idea of GLTL is roughly to restrict the
period of validity of the temporal operators to bounded
windows---similar to the bounded semantics of LTL~\citep{pnueli-specification}. To this
end, GLTL introduces operators of the form of $\eventually_\mu b$ with
the atomic proposition $b$, which is interpreted as ``$b$ eventually
holds within $k$ time steps where $k$ is a random variable following a
geometric distribution with parameter $\mu$.'' Similar semantics
stochastically restricting the window of validity for other temporal
operators are also introduced.

This kind of geometric decay fits very nicely with MDPs for a few
reasons. It can be viewed as a generalization of reward discounting,
which is already present in many MDP models. It also avoids
unnecessarily expanding the specification state space by only
requiring extra states to represent events and not simply the passage
of time.

Using $G1(\mu)$ to represent the geometric distribution with parameter
$\mu$, the temporal operators are:
\begin{itemize}
\item $\eventually_\mu p$: $p$ is achieved in the next $k$ steps,
  $k\sim G1(\mu)$.
\item $\always_\mu q$: $q$ holds for the next $k$ steps,
$k\sim G1(\mu)$.
\item $q \until_\mu q$: $q$ must hold at least until $p$ becomes
  true, which itself must be achieved in the next $k$ steps, $k\sim
  G1(\mu)$.
\end{itemize}

\begin{figure*}
\vskip 0.2in
\begin{center}
\centerline{\includegraphics[width=0.50\textwidth]{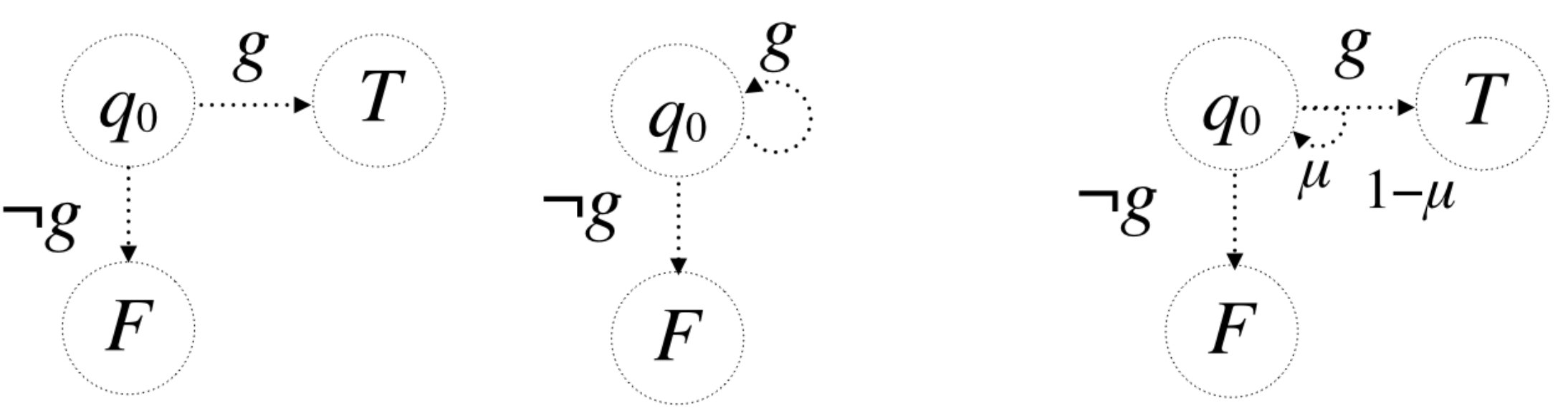}
\includegraphics[width=0.25\textwidth]{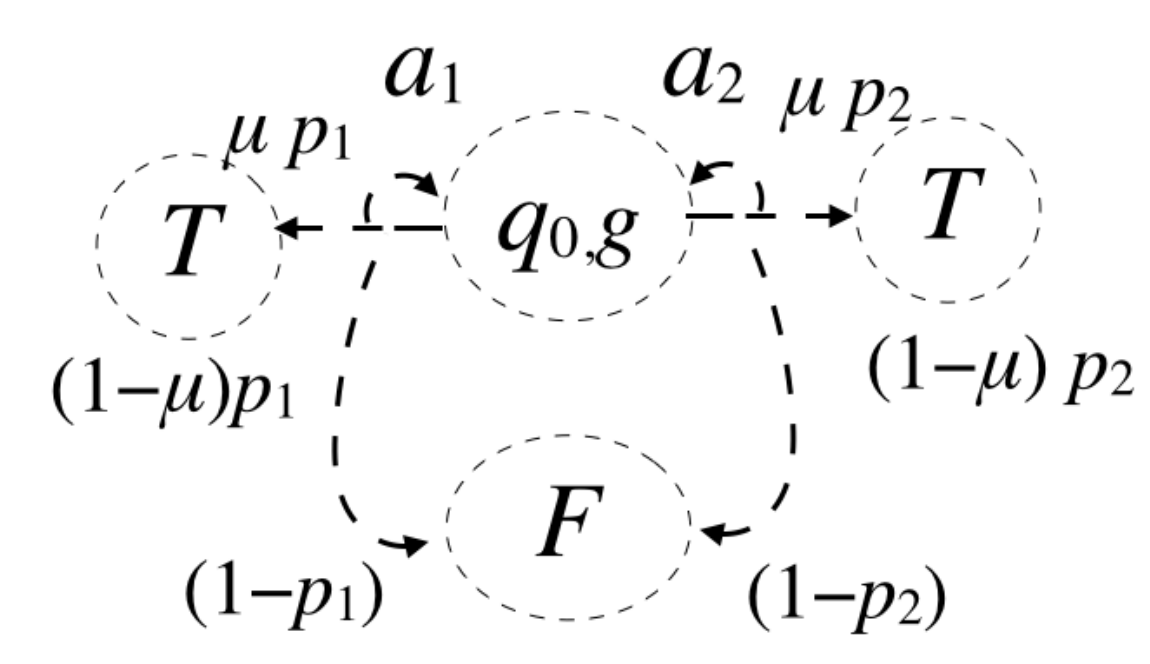}}
\caption{{\it First}: The specification MDP representation of the LTL formula
  $b$. {\it Second}: The specification MDP representation of the LTL formula
  $\always b$. {\it Third}: The specification MDP representation of
  the GLTL formula $\always_\mu b$. {\it Fourth}: The composition of the specification MDP representation of
  the GLTL formula $\always_\mu b$ with the MDP from Figure~\ref{f:unlearn}.}
\label{f:always}
\end{center}
\vskip -0.2in
\end{figure*}

Returning to our earlier example from Figure~\ref{f:unlearn},
evaluating the probability of satisfaction for $\always g$ requires
infinite precision in the learned transition probabilities in the
environment. Consider instead evaluating $\always_{\mu} g$ in this
environment. An encoding of the specification for this example is
shown in Figure~\ref{f:always}~(\emph{Third}). We call it a
\emph{specification MDP}, as it specifies the task using states
(derived from the formula), actions (representing conditions), and
probabilities (capturing the stochasticity of operator
expiration). This example says that, from the initial state $q_0$,
encountering any state where $g$ is not true results in immediately
failing the specification. In contrast, encountering any state where
$g$ is true results in either continued evaluation (with probability
$\mu$) or success (with probability $1-\mu$). Success represents the
idea that the temporal window in which $g$ must hold true has expired
without $g$ being violated.

Composing these two MDPs leads to the composite MDP in
Figure~\ref{f:always}~(\emph{Fourth}). The true satisfaction
probability for action $a_i$ is $\frac{1-\mu}{(1-\mu + p_i)}$. Thus,
if $\mu=.9$, the dependence of this value on $\epsilon$ is
$\frac{.1}{.1 + \epsilon}$, which is well behaved for all values of
$\epsilon$. The sensitivity of the computed satisfaction probability
has a maximum $1/(1-\mu)^2$ dependence on the accuracy of the estimate
of $\epsilon$. Thus, GLTL is considerably more friendly to learning
than is LTL.

Returning to the MDP example in Figure~\ref{f:nopenalty}, we find that
GLTL is also more expressive than rewards. The GLTL formula $\neg q
\until_{\mu} p$ can be translated to a specification MDP.
Essentially, the idea is that encountering a bad
state ($q$) even once or running out of time
results in specification
failure. Maximizing the satisfaction of this GLTL formula results in taking action $a_1$
regardless of the value of $p$. That is, it is an
environment-independent specification of the task.

The reason the GLTL formulation is able to succeed where standard
rewards fail is that the GLTL formula results in an augmentation of
the state space so that the reward function can depend on whether a
bad state has yet been encountered. On the first encounter, a penalty
can be issued. After the first encounter, no additional penalty is
added. By composing the environment MDP with this bit of internal
memory, the task can be expressed provably correctly and in an
environment-independent way.

\section{Related Work}

Discounting has been used in previous temporal models. In quantitative temporal
logic, it gives more weight to the satisfaction of a logic property in
the near future than the far
future. \cite{de2003discounting,de2004model} augment computation tree
logic (CTL) with discounting and develop fixpoint-based algorithms for
checking such properties for probabilistic systems and games.
\cite{almagor2014discounting} explicitly refine the ``eventually''
operator of LTL to a \emph{discounting operator} such that the longer
it takes to fulfill the task the smaller the value of
satisfaction. Further, they show that discounted LTL is more
expressive than discounted CTL. They use both discounted until and
undiscounted until for expressing traditional eventually as well as
its discounted version. However, algorithms for model checking and
synthesis discounted LTL for probabilistic systems and games are yet
to be developed.

LTL has been used extensively in robotics domains.  Work on the
trustworthiness of autonomous robots, automated verification and
synthesis with provable correctness with respect to temporal
logic-based specifications in motion, task, and mission planning have
attracted considerable attention recently. The results include
open-loop and reactive control of deterministic, stochastic or
non-deterministic finite-state models as well as continuous state
models through appropriate finite-state
abstractions~\citep{itac,gazit09, liu2013synthesis,eric-cdc, ding,
  calinacc, ram}. While temporal logic had initially focused on
reasoning about temporal and logical relations, its dialects with
probabilistic modalities have been used increasingly for robotics
applications~\citep{modelcheckingbook, DeAlfaro, prism1}.

\section{Generating Specification MDPs}

Similar to LTL, GLTL formulas are built from a set of atomic propositions $AP$, Boolean operators $\wedge$ (conjunction), $\neg$ (negation) and temporal operator $\mathcal{U}_{\mu}$ ($\mu$-until).  Useful operators such as $\vee$ (disjunction), $\eventually_{\mu}$ ($\mu$-eventually) and $\always_{\mu}$ ($\mu$-always) can be derived from these basic operators.

GLTL formulas can be converted to the corresponding specification MDPs recursively, with the operator precedence listed in descending order in Table~\ref{table:operator_precedence}. Operators of the same precedence are read from right to left. For example, $\always_{\mu_1} \eventually_{\mu_2} \varphi = (\always_{\mu_1} (\eventually_{\mu_2} \varphi))$, $\varphi_1 \mathcal{U}_{\mu_1} \varphi_2 \mathcal{U}_{\mu_2} \varphi_3 = (\varphi_1 \mathcal{U}_{\mu_1} (\varphi_2 \mathcal{U}_{\mu_2} \varphi_3))$.

\begin{table}
	\caption{Operator precedence in specification MDP construction.}
	\label{table:operator_precedence}
	\centering
	\begin{tabular}{| c | c | c |} \hline
	Precedence & Operator & \# of Operands \\ \hline
	 1 & not, $\neg$ & 1 \\ \hline
	 \multirow{3}{*}{2} & $\mu$-always, $\always_{\mu}$ & 1 \\
	 & $\mu$-eventually, $\eventually_{\mu}$ & 1 \\
	 & $\mu$-until, $\mathcal{U}_{\mu}$ & 2 \\ \hline
	 3 & and, $\wedge$ & 2 \\ \hline
	 4 & or, $\vee$ & 2 \\ \hline
	\end{tabular}
\end{table}

Assume $\varphi, \varphi_1, \varphi_2$ are GLTL formulas in the following discussion.
\begin{itemize}
\item $b$, where $b \in AP$ is an atomic proposition: A specification
  MDP $M_b = (\{s^{ini}, \acc, \rej\}, \{a\}, T, R)$ for $b$ can be
  constructed such that, if $p$ holds at $s^{ini}$, the transition
  $(s^{ini}, a, \acc)$ is taken with probability 1; otherwise, the
  transition $(s^{ini}, a, \rej)$ is taken with probability 1.
\item $\neg \varphi$: A specification MDP $M_{\neg \varphi}$ can be constructed from a specification MDP $M_{\varphi}$ by swapping the terminal states $\acc$ and $\rej$.
\item $\varphi_1 \wedge \varphi_2$: A specification MDP $M_{\varphi_1 \wedge \varphi_2} = (S, A, T, R)$ can be constructed from specification MDPs $M_{\varphi_1} = (S_1, A_1, T_1, R_1)$ and $M_{\varphi_2} = (S_2, A_2, T_2, R_2)$ such that (1) $S = (S_1 \backslash \{\rej_1\}) \times (S_2 \backslash \{\rej_2\}) \bigcup \{\rej\}$, and the accepting state is $\acc = (\acc_1, \acc_2)$; (2) $A = A_1 \times A_2$; (3) for all transitions $(s_1, a_1, s'_1)$ of $M_{\varphi_1}$ and $(s_2, a_2, s'_2)$ of $M_{\varphi_2}$, if either $s'_1 = \rej_1$ or $s'_2 = \rej_2$, let $T((s_1, s_2), (a_1, a_2), \rej) = T_1(s_1, a_1, s'_1) T_2(s_2, a_2, s'_2)$; otherwise, $T((s_1, s_2), (a_1, a_2), (s'_1, s'_2)) = T_1(s_1, a_1, s'_1) T_2(s_2, a_2, s'_2)$.
\item $\varphi_1 \vee \varphi_2 = \neg (\neg \varphi_1 \wedge \neg \varphi_2)$.
\item $\varphi_1 \mathcal{U}_{\mu} \varphi_2$: The operator $\mu$-until has two operands, $\varphi_1$ and $\varphi_2$, which generate specification MDPs $M_{\varphi_1} = (S_1, A_1, T_1, R_1)$ and $M_{\varphi_2} = (S_2, A_2, T_2, R_2)$. The new specification MDP $M_{\varphi_1 \mathcal{U}_{\mu} \varphi_2} = (S, A, T, R)$ is constructed from $M_{\varphi_1}$ and $M_{\varphi_2}$: $S = (S_1 \backslash \{\acc_1, \rej_1\}) \times (S_2 \backslash \{\acc_2, \rej_2\})\bigcup \{\acc, \rej\}$, where $\acc$ and $\rej$ are the accepting and rejecting state, respectively, and $s^{ini} = (s^{ini}_1, s^{ini}_2) \in S$ is the initial state; $A = A_1 \times A_2$; for all $s = (s_1, s_2) \in S \backslash \{\acc, \rej\}$, $a = (a_1, a_2) \in A$, $s'_1 \in S_1$ and $s'_2 \in S_2$, if $T_1(s_1,a_1,s'_1) > 0$ and $T_2(s_2, a_2, s'_2) > 0$, a transition $(s,a,s')$ is added to $M_{\varphi_1 \mathcal{U}_{\mu} \varphi_2}$ with probability $T(s,a,s')$ as specified in Table~\ref{table:until}.

Here are some intuitions behind the construction of $T$. The formula
$\varphi_1 \mathcal{U}_{\mu} \varphi_2$ means that, within some
stochastically decided time period $k$, we would like to successfully
implement task $\varphi_2$ in at most $k$ steps without ever failing
in task $\varphi_1$. If we observe a success in $M_{\varphi_2}$ (that
is, the specification reaches $\acc_2$) before $\varphi_1$ fails (that
is  the sepcification reaches $\rej_1$), $M_{\varphi_1
  \mathcal{U}_{\mu} \varphi_2}$ goes to state $\acc$ for sure; if we
observe a failure in $M_{\varphi_1}$ (that, the specifcation reaches
$\rej_1$) before succeeding in $M_{\varphi_2}$ (that is, the
specification reaches $\acc_2$), $M_{\varphi_1 \mathcal{U}_{\mu}
  \varphi_2}$ goes to state $\rej$ for sure. In all other cases,
$M_{\varphi_1 \mathcal{U}_{\mu} \varphi_2}$ primarily keeps track of
the transitions in $M_{\varphi_1}$ and $M_{\varphi_2}$, with a tiny
probability of failing immediately, which corresponds to the operator expiring.

\item $\eventually_{\mu} \varphi_2$: As in the semantics of LTL, $\mu$-eventually $\eventually_{\mu} \varphi_2 = \text{True } \mathcal{U}_{\mu} \varphi_2$. Hence, given a specification MDP $M_{\varphi_2} = (S_2, A_2, T_2, R_2)$ for $\varphi_2$, we can construct a specification MDP $M_{\eventually_{\mu} \varphi_2} = (S, A, T, R)$ for $\eventually_{\mu} \varphi_2$: $S = S_2$, $s^{ini} = s^{ini}_2$, $\acc = \acc_2$, $\rej = \rej_2$; $A = A_2$; transitions of $M_{\eventually_{\mu} \varphi_2}$ are modified from those of $M_{\varphi_2}$ as in Table~\ref{table:eventually}.
Informally, $\eventually_{\mu} \varphi_2$ is satisfied if we succeed in task $\varphi_2$ within the stochastic observation time period.

\item $\always_{\mu} \varphi_2$: $\mu$-always $\varphi_2$ is equivalent to $\neg \eventually_{\mu} \neg \varphi_2 = \neg (\eventually_{\mu} (\neg \varphi_2))$. In other words, $\always_{\mu} \varphi_2$ is satisfied if we did not witness a failure of $\varphi_2$ within the stochastic observation time period. The transitions of a specification MDP $M_{\always_{\mu} \varphi_2}$ can be constructed from Table~\ref{table:eventually}, or directly from Table~\ref{table:always}.

\end{itemize}

\begin{table}
  \caption{Transition $(s,a,s')$ in $M_{\varphi_1 \mathcal{U}_{\mu}
      \varphi_2}$ constructed from a transition $(s_1, a_1, s'_1)$ in
    $M_{\varphi_1}$ and a transition $(s_2, a_2, s'_2)$ in
    $M_{\varphi_2}$. Here, $p(s' | s'_1, s'_2) = \frac{T(s,a,s')}{T_1(s_1, a_1, s'_1) T_2(s_2, a_2, s'_2)}$. 
	That is, to get the transition probability, multiple the $p$
        column by the corresponding $T_1$ and $T_2$ transition probabilities.}
  \label{table:until}
  \vspace{2ex}
  \centering
  \begin{adjustbox}{max width=\linewidth}
  \begin{tabular}{| c | c | c | c |} \hline
	  $s'_1$ & $s'_2$ & $s'$ & $p(s'|s'_1, s'_2)$\\ \hline \hline
	  $\acc_1$ & $\acc_2$ & $\acc$ & 1\\  \hline
	  \multirow{2}{*}{$\acc_1$} & \multirow{2}{*}{$\rej_2$} & $(s'_1, s'_2)$ & $1 - \mu$ \\ \cline{3-4}
	  & & $\rej$ & $\mu$ \\ \hline
	  \multirow{2}{*}{$\acc_1$} & \multirow{2}{*}{$S_2 \backslash \{\acc_2, \rej_2\}$} & $(s^{ini}_1, s'_2)$ & $1 - \mu$ \\ \cline{3-4}
	  & & $\rej$ & $\mu$ \\ \hline
	  $\rej_1$ & $\acc_2$ & $\acc$ & 1 \\ \hline
	  $\rej_1$ & $S_2 \backslash \{\acc_2\}$ & $\rej$ & 1 \\ \hline
	  $S_1 \backslash \{\acc_1, \rej_1\}$ & $\acc_2$ & $\acc$ & 1 \\ \hline
	  \multirow{2}{*}{$S_1 \backslash \{\acc_1, \rej_1\}$} & \multirow{2}{*}{$\rej_2$} & $(s'_1, s^{ini}_2)$ & $1-\mu$ \\ \cline{3-4}
	  & & $\rej$ & $\mu$ \\ \hline
	  \multirow{2}{*}{$S_1 \backslash \{\acc_1, \rej_1\}$} & \multirow{2}{*}{$S_2 \backslash \{\acc_2, \rej_2\}$} & $(s'_1, s'_2)$ & $1-\mu$ \\ \cline{3-4}
	  & & $\rej$ & $\mu$ \\ \hline
  \end{tabular}
  \end{adjustbox}
\end{table}

\begin{table}
  \caption{Transition $(s,a,s')$ in $M_{\eventually_{\mu} \varphi_2}$
    constructed from a transition $(s_2, a_2, s'_2)$ in
    $M_{\varphi_2}$. As above, $p(s' | s'_2) = \frac{T(s,a,s')}{T_2(s_2, a_2, s'_2)}$. }
  \label{table:eventually}
  \vspace{2ex}
  \centering
  \begin{adjustbox}{max width=\linewidth}
  \begin{tabular}{| c | c | c |} \hline
	  $s'_2$ & $s'$ & $p(s'|s'_2)$\\ \hline \hline
	  $\acc_2$ & $\acc_2$ & 1\\  \hline
	  \multirow{2}{*}{$\rej_2$} & $s^{ini}_2$ & $1 - \mu$ \\ \cline{2-3}
	  & $\rej_2$ & $\mu$ \\ \hline
	  \multirow{2}{*}{$S_2 \backslash \{\acc_2, \rej_2\}$} & $s'_2$ & $1 - \mu$ \\ \cline{2-3}
	  & $\rej_2$ & $\mu$ \\ \hline
  \end{tabular}
  \end{adjustbox}
\end{table}

\begin{table}
  \caption{Transition $(s,a,s')$ in $M_{\always_{\mu} \varphi_2}$
    constructed from a transition $(s_2, a_2, s'_2)$ in
    $M_{\varphi_2}$. As above, $p(s' | s'_2) = \frac{T(s,a,s')}{T_2(s_2, a_2, s'_2)}$. }
  \label{table:always}
  \vspace{2ex}
  \centering
  \begin{adjustbox}{max width=\linewidth}
  \begin{tabular}{| c | c | c |} \hline
	  $s'_2$ & $s'$ & $p(s'|s'_2)$\\ \hline \hline
	  \multirow{2}{*}{$\acc_2$} & $s^{ini}_2$ & $1 - \mu$ \\ \cline{2-3}
	  & $\acc_2$ & $\mu$ \\ \hline
	  $\rej_2$ & $\rej_2$ & 1\\  \hline
	  \multirow{2}{*}{$S_2 \backslash \{\acc_2, \rej_2\}$} & $s'_2$ & $1 - \mu$ \\ \cline{2-3}
	  & $\acc_2$ & $\mu$ \\ \hline
  \end{tabular}
  \end{adjustbox}
\end{table}

Using the transitions as described, a given GLTL formula can be
converted into a specification MDP. To satisfy the specification in a
given environment, a joint MDP is created as follows:
\begin{enumerate}
\item Take the cross product of the MDP representing the environment
  and the specification MDP.
\item Any state that corresponds to an accepting or rejecting state of
  the specification MDP becomes a sink state. However, the accepting
  states also include a reward of $+1$.
\item The resulting MDP is solved to create a policy.
\end{enumerate}

The resulting policy is one that maximizes the probability of
satisfying the given formula where the random events are both the
transitions in the environment and the stochastic transitions in the
specification MDP. Such policies tend to prefer satisfying formulas
quickly, as that increases the chance of successful completion before
operators expire.

\section{Example Domain}

\newcommand{\blue}{\mbox{\rm blue}}
\newcommand{\red}{\mbox{\rm red}}
\newcommand{\green}{\mbox{\rm green}}

Consider the following formula:
$$(\neg\blue \until_\mu \red) \wedge (\eventually_\mu (\red \wedge \eventually_\mu \green)).$$
It specifies a task of reaching a red state without encountering a
blue state and, once a red state is reached, going to a green state.

\begin{figure}
\vskip 0.2in
\begin{center}
\centerline{\includegraphics[width=0.4\textwidth]{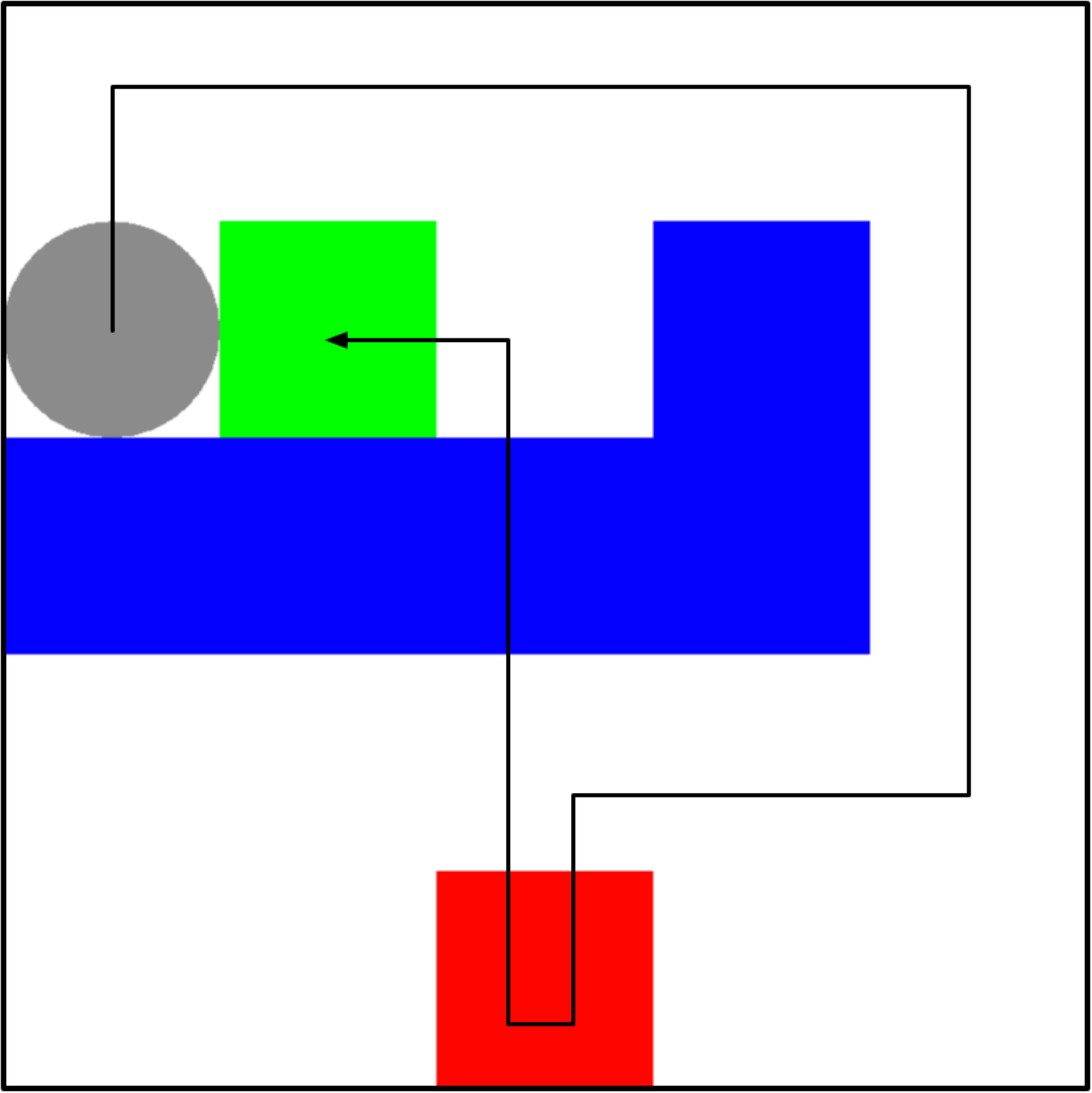}}
\caption{The optimal path in a slightly more complex  grid world.}
\label{f:grid1}
\end{center}
\vskip -0.2in
\end{figure}

Figure~\ref{f:grid1} illustrates a grid world environment in which
this task can be carried out. It consists of different colored grid
cells. The agent can move to one of the four adject cells to its
current position with a north, south, east, or west action. However,
selecting an action for one direction has a 0.02 probability of moving in the one
of the three other directions. This stochastic movement causes the
agent to keep its distance from dangerous grid cells that could result
in task failure, whenever possible.  The solid line in the figure
traces the path of the optimal policy of following this specification
in the grid. As can be seen, the agent moves to red and then
green. Note that this behavior can be very difficult to encode in a
standard reward function as both green and red need to be given
positive reward and therefore either would be a sensible place for the
agent to stop.

Figure~\ref{f:grid1} illustrates a grid world environment in which the
blue cells create a partial barrier between the red and green
cells. As a result of the ``until'' in the specification, the agent
goes around the blue wall to get to the red cell. However, since the
prohibition against blue cells is lifted once the red cell is reached,
it goes directly through the barrier to reach green.

\begin{figure}
\vskip 0.2in
\begin{center}
\centerline{\includegraphics[width=0.4\textwidth]{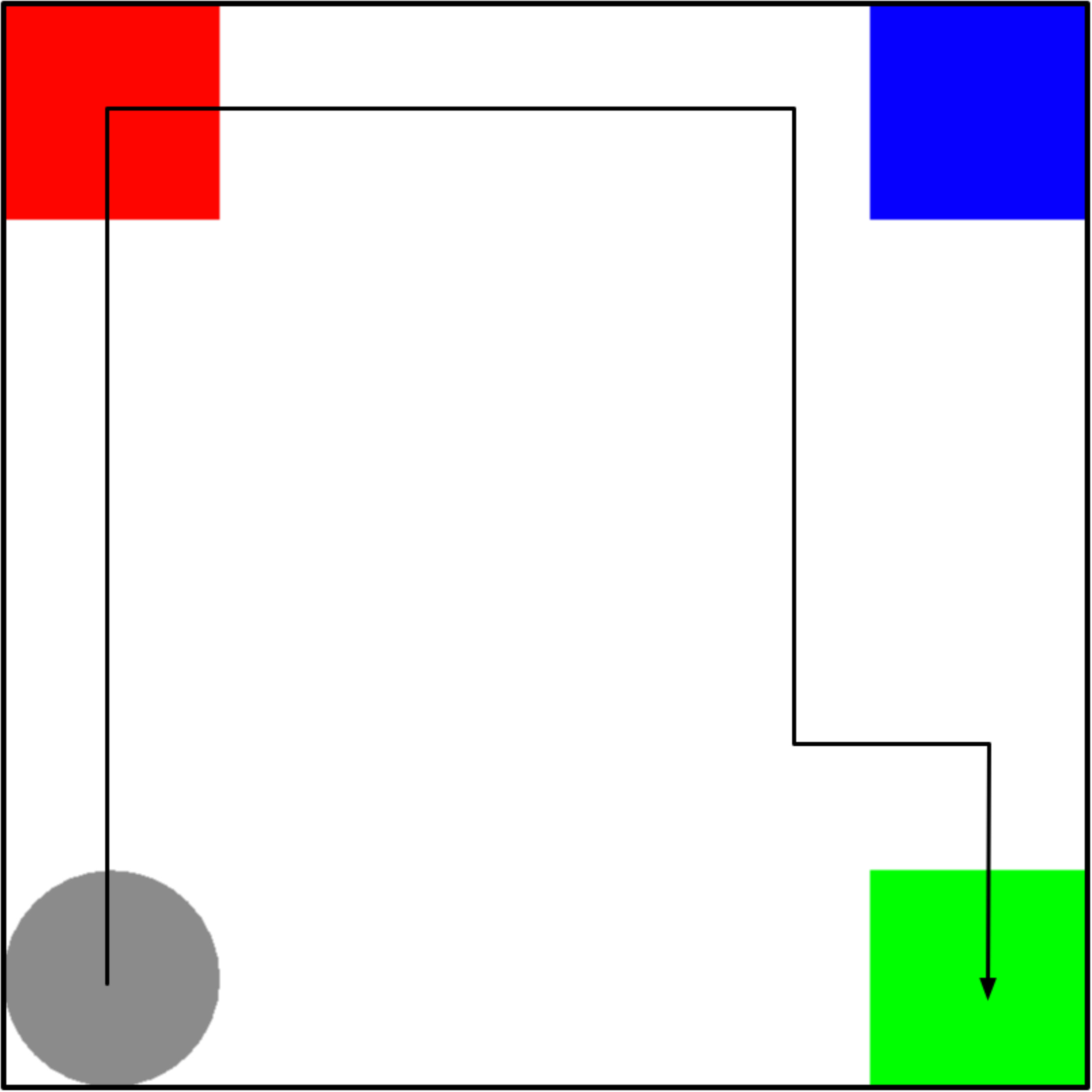}}
\caption{The optimal path in a grid world.}
\label{f:grid3}
\end{center}
\vskip -0.2in
\end{figure}

These 25-state environments become 98-state MDPs when combined with
the specification MDP.

\section{Conclusion}

In contrast to standard MDP reward functions, we have provided an
environment-independent specification for tasks. We have shown that
this specification language can capture standard tasks used in the MDP
community and that it can be automatically incorporated into an
environment MDP to create a fixed MDP to solve. Maximizing reward in
this resulting MDP maximizes the probability of satisfying the task
specification.

Future work includes inverse reinforcement learning of task
specifications and techniques for accelerating planning.


\bibliography{../mlittman,ufuk}
\bibliographystyle{icml2015}

\end{document}